\documentclass[runningheads]{llncs}

\usepackage{eccv}

\usepackage{eccvabbrv}

\usepackage{graphicx}
\usepackage{booktabs}

\usepackage[accsupp]{axessibility}  
\usepackage{makecell}
\usepackage{xspace}
\usepackage{indentfirst}
\usepackage{listings}
\usepackage{multirow}            
\usepackage{enumerate}
\usepackage{xcolor}
\usepackage{longtable}

\usepackage[pagebackref,breaklinks,colorlinks,citecolor=eccvblue]{hyperref}
\usepackage{hyperref}

\usepackage{orcidlink}

\newcommand\blfootnote[1]{%
\begingroup
\renewcommand\thefootnote{}\footnote{#1}%
\addtocounter{footnote}{-1}%
\endgroup
}

\definecolor{red}{rgb}{1.0,0,0}
\definecolor{codegreen}{rgb}{0,0.6,0}
\definecolor{codegray}{rgb}{0.5,0.5,0.5}
\definecolor{codepurple}{rgb}{0.58,0,0.82}
\definecolor{backcolour}{rgb}{0.95,0.95,0.92}
\definecolor{lightblue}{RGB}{62,129,183}

\lstdefinestyle{mystyle}{
  backgroundcolor=\color{backcolour}, commentstyle=\color{codegreen},
  keywordstyle=\color{magenta},
  numberstyle=\tiny\color{codegray},
  stringstyle=\color{codepurple},
  basicstyle=\ttfamily\footnotesize,
  breakatwhitespace=false,         
  breaklines=true,                 
  captionpos=b,                    
  keepspaces=true,                 
  numbers=left,                    
  numbersep=5pt,                  
  showspaces=false,                
  showstringspaces=false,
  showtabs=false,                  
  tabsize=2
}

\begin{document}

\title{Lite-SAM Is Actually What You Need for Segment Everything} 



\author{
Jianhai Fu\inst{1,*}\orcidlink{0009-0009-2819-3717} \and
Yuanjie Yu\inst{1,2,*}\orcidlink{0009-0006-1245-3316} \and
Ningchuan Li\inst{1,\dagger}\orcidlink{0009-0005-2852-1312} \and
Yi Zhang\inst{1}\orcidlink{0009-0002-5631-5637} \and \\
Qichao Chen\inst{1}\orcidlink{0009-0008-7580-0068} \and
Jianping Xiong\inst{1}\orcidlink{0000-0001-9564-1860} \and
Jun Yin\inst{2}\orcidlink{0009-0007-1905-5377}  \and
Zhiyu Xiang\inst{2,\dagger}\orcidlink{0000-0002-3329-7037}
}

\institute{Zhejiang Dahua Technology Co., Ltd., Hangzhou, China
\and
Zhejiang University, Hangzhou, China}

\authorrunning{Fu et al.}


\maketitle

\blfootnote{ $^{*}$Equal contribution.}
\blfootnote{ $^{\dagger}$Corresponding author.}

\begin{abstract}
The Segment Anything model (SAM) has brought significant changes 
to the segmentation field with its superior performance, 
but its extensive computational resource requirements remain a limiting factor.
Many works, such as MobileSAM, Edge-SAM, and MobileSAM-v2, 
have explored lightweight solutions.
However, their use of traditional Grid Search sampling strategies 
or two-stage concatenation methods, which do not allow for end-to-end training, 
severely limit the performance of segment everything (SegEvery).

This paper introduces Lite-SAM, an efficient end-to-end solution 
for the SegEvery task designed to reduce computational costs and redundancy.
Lite-SAM is composed of four main components: 
a streamlined CNN-Transformer hybrid encoder (LiteViT), 
an automated prompt proposal network (AutoPPN), a traditional prompt encoder, and a mask decoder. 
All these components are integrated within the SAM framework.
Our LiteViT, a high-performance lightweight backbone network, has only 1.16M parameters,
which is a 23$\%$ reduction compared to the lightest existing backbone network Shufflenet. 
We also introduce AutoPPN, an innovative end-to-end method for prompt boxes and points generation. 
This is an improvement over traditional grid search sampling methods, 
and its unique design allows for easy integration into any SAM series algorithm, extending its usability.

we have thoroughly benchmarked Lite-SAM across a plethora of both public and private datasets.
The evaluation encompassed a broad spectrum of universal metrics, 
including the number of parameters, SegEvery execution time, and accuracy. 
The findings reveal that Lite-SAM, operating with a lean 4.2M parameters, significantly outpaces its counterparts,
demonstrating performance improvements of 43x, 31x, 20x, 21x, and 1.6x 
over SAM, MobileSAM, Edge-SAM, EfficientViT-SAM, and MobileSAM-v2 respectively, 
all the while maintaining competitive accuracy. 
This underscores Lite-SAM's prowess in achieving an optimal equilibrium between performance and precision, 
thereby setting a new state-of-the-art(SOTA) benchmark in the domain.

\end{abstract}

\section{Introduction}
\label{sec:intro}

\begin{figure}[!ht]
	\centering
	\includegraphics[width=1\linewidth]{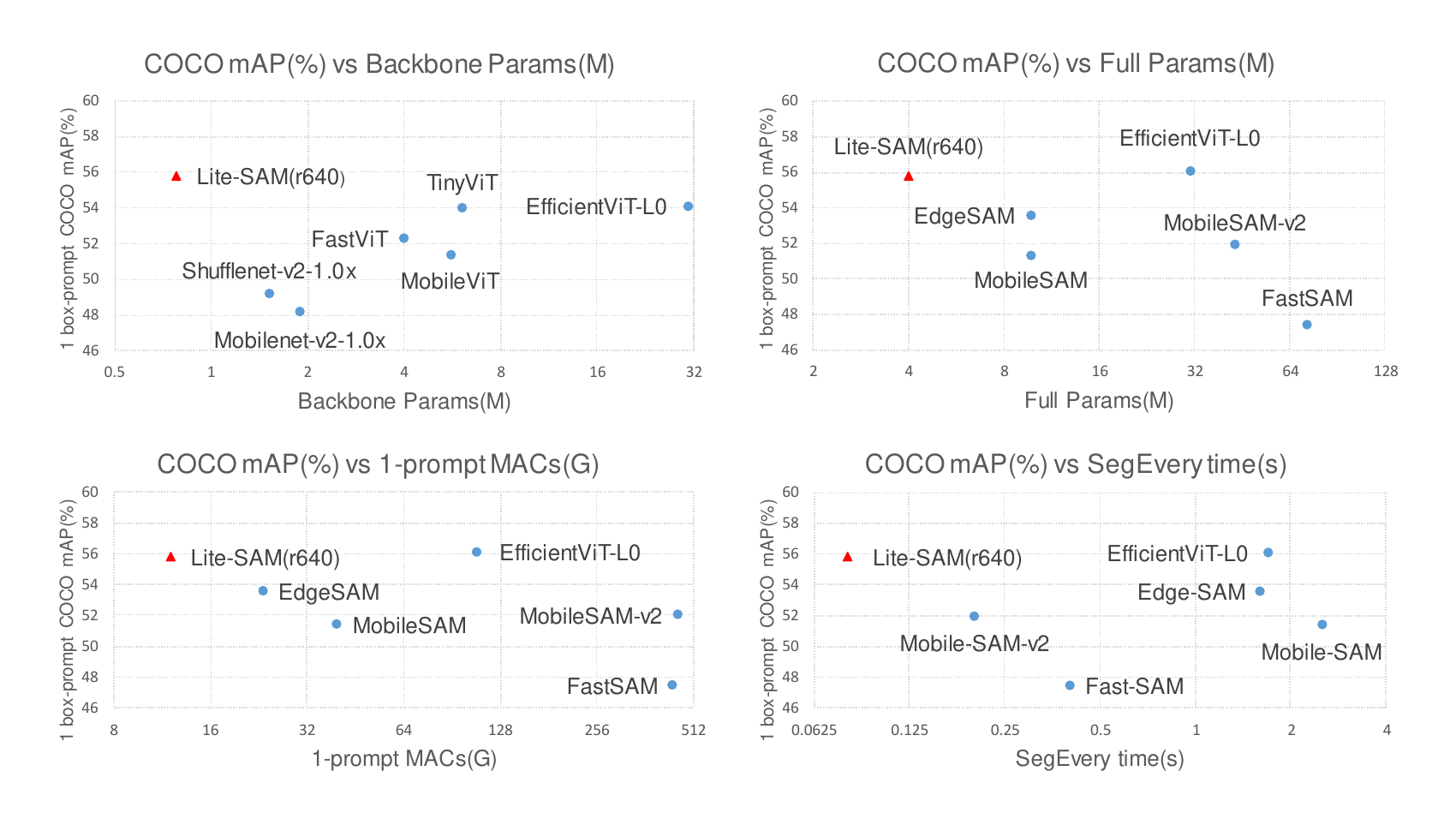}	
	\caption{
     The proposed Lite-SAM achieves SOTA performance in terms of Backbone Parameters (top left), 
	Full Parameters (top right), Multiply-Accumulate Operations (bottom left), 
	and SegEvery time (bottom right) tasks while maintaining computational efficiency. 
	The metrics were evaluated on the zero-shot learning of the COCO dataset.
     Note that the comparison of backbone parameters is made against 
     lightweight network structures (params $\leq$ 40M), with MAE not falling within this scope. 
	}
	\label{fig:SOTA_overview}
\end{figure}

Zhang et al. \cite{zhang2023one} have made a remarkable leap in the field of NLP,
resulting in a significant breakthrough in generative AI 
(AIGC, also known as Artificial Intelligence Generated Content) \cite{zhang2023complete}.
This breakthrough has largely been enabled by the GPT-series models \cite{radford2018improving,brown2020language},
which are foundation models \cite{bommasani2021opportunities} trained on extensive text datasets.
Capitalizing on the success of these foundation models in NLP, multiple studies \cite{zhang2022does,he2020momentum,qiao2023mp}
have explored the integration of image encoders and text encoders via contrastive learning \cite{zhang2022dual}.

The Meta Research team has recently introduced an ambitious program known as ``segment everything''
project called SAM \cite{kirillov2023segment}. SAM represents a crucial advancement for vision framework,  
drawing parallels to the impact of GPT in NLP.
It comprises two key components: a ViT-based image encoder and a prompt-guided mask decoder,
which work in conjunction. 
SAM is designed to handle two segmentation tasks: SegAny and SegEvery.
Both tasks involve class-agnostic mask segmentation but differ in their objectives. 
SegAny uses a specific prompt, such as a point or box, 
to isolate and segment a particular item of interest within an image. 
In contrast, SegEvery's objective is to delineate all discernible subjects in the image.

Chaoning Zhang et al. \cite{zhang2023faster} proposed a ``decoupled distillation''
aimed at distilling the ViT-H decoder of SAM \cite{kirillov2023segment}, 
which yielded a more efficient lightweight encoder that could integrate with SAM's decoder.
However, this algorithm model lacks robustness in platform adaptation 
and exhibits considerable accuracy loss during such translations, 
rendering it less suitable for deployment on mobile devices. 
Zhao et al. \cite{zhao2023fast} introduced the Fast-SAM model, built upon YOLOv8 \cite{yolov8_ultralytics1}, 
that demonstrates remarkable segmentation capabilities. 
Its main limitation, however, is the absence of a full range of interactive modalities, 
notably lacking in dedicated box and point functionalities.
Li et al. \cite{li2023semantic} engineered SemanticSAM, 
a model that enhances the segmentation and recognition versatility of images across varying scales. 
It is imperative to highlight, though, that its substantial number of parameters contributes to longer inference times. 
Han Cai et al. \cite{caiefficientvit} presented EfficientViT, introduced a novel lightweight algorithm 
called which achieved promising results. 
Chong Zhou et al. proposed Edge-SAM \cite{zhou2023edgesam}, 
an algorithm that accomplishes real-time execution for the SegAny task on an iPhone. 
All the aforementioned methods \cite{zhang2023faster,kirillov2023segment,zhao2023fast,li2023semantic,caiefficientvit,zhou2023edgesam,xiong2023efficientsam}
are all evaluated for SegAny; however, the SegEvery continues to be highly time-demanding.

MobileSAM-v2 \cite{zhang2023mobilesamv2} proposed an innovative training approach for YOLOv8 \cite{yolov8_ultralytics1} 
that uses pre-generated prompts (Object-Aware Prompt Sampling) in place of the traditional Gridsearch sampling strategy, 
enhancing the efficiency of the SegEvery process. Despite this improvement, 
this approach necessitates the use of separate models, which is considered a stopgap measure. 
Due to YOLOv8's inherent inference and training demands, the overall time savings may be limited.

In order to address the aforementioned issues, our contributions can be summarized as follows:
\begin{itemize}
\item We Introduced LiteViT, a lightweight CNN-Transformer encoder, 
enhancing accuracy with reduced parameters, ideal for limited computational environments.

\item The development of AutoPPN, an automated prompt proposal network, 
improving efficiency over grid search methods and integrating easily with SAM series algorithms.

\item Validated Lite-SAM's performance through experiments, as depicted in \cref{fig:SOTA_overview}, 
showing accelerated results on SegEvery while preserving accuracy.

\end{itemize}

\section{Related Works}
\label{sec:related}

\subsection{Segment Anything}
In the evolving field of image segmentation, the SAM~\cite{kirillov2023segment} stands out as a significant progress. 
Its groundbreaking training methodology and exceptional performance on extensive visual datasets distinguish it.
SAM is particularly adept in class-agnostic segmentation and shows impressive efficacy in zero-shot scenarios. 

In addition to the work on lightweight versions of Segment Anything and its variants 
mentioned in ~\cite{zhang2023faster,kirillov2023segment,zhao2023fast,li2023semantic,caiefficientvit,zhou2023edgesam,xiong2023efficientsam,zhang2023mobilesamv2}, 
a series of works combining SAM with various downstream tasks have also achieved impressive results.

Grounded SAM~\cite{ren2024grounded} integrates Grounding DINO's open-set detection with SAM, 
enabling text-guided detection and segmentation in images. 
SegGPT~\cite{wang2023seggptsegmentingcontext} standardizes diverse segmentation data into a single image format, 
excelling in segmenting both in-domain and out-of-domain subjects with strong performance. 
Zou et al.~\cite{zou2023segment} presented SEEM, featuring a versatile decoding mechanism for various segmentation tasks, 
aiming to create a universal interface akin to large language models. 
Inpaint Anything ~\cite{yu2023inpaintanythingsegmentmeets} introduces a novel ``click and fill'' method 
for mask-free image inpainting, blending SAM models with AIGC to create an efficient and user-friendly solution for inpainting tasks.
SAM3D~\cite{yang2023sam3dsegment3dscenes} advances 3D perception by mapping 2D segmentation to 3D spaces. 
It enables 3D point cloud mask prediction using RGB images with the SAM model, eliminating the need for additional training or fine-tuning.

As a multipurpose foundational model,
SAM has greatly enhanced interactive segmentation techniques and demonstrated remarkable flexibility across diverse segmentation tasks.
Its contributions has notably expanded the horizons for applications in open-world image understanding.
However, a noteworthy limitation of SAM is its constrained real-time processing capabilities, 
which poses obstacles for time-sensitive applications.

\subsection{Lightweight ViT and CNN}
Historically, mobile vision applications have heavily relied on lightweight Convolutional Neural Networks (CNNs) like
 MobileNet~\cite{howard2017mobilenets} and ShuffleNet~\cite{zhang2018shufflenet,ma2018shufflenet}.
The MobileNet series~\cite{sandler2018mobilenetv2,howard2019searching} was pioneering in its segmentation of 
convolution blocks into depth-wise and point-wise convolutions,
significantly reducing model size and computational demand. 
The emergence of Vision Transformers (ViTs)~\cite{dosovitskiy2020image} has spurred efforts to streamline these architectures,
resulting in more compact and efficient models such as Deit-Small (Deit-S) and Deit-Tiny (Deit-T)~\cite{touvron2021training}.
MobileViT~\cite{mehta2021mobilevit} fused ViTs with conventional convolutions, outperforming MobileNet-v2~\cite{sandler2018mobilenetv2} 
by focusing on improved local feature extraction, a forte of CNNs. 
The trend toward computational economy is further advanced by subsequent models, 
including EfficientFormer~\cite{li2022efficientformer}, EfficientViT~\cite{liu2023efficientvit}, 
Next-ViT~\cite{li2022next}, TinyViT~\cite{wu2022tinyvit}, and FastViT~\cite{vasu2023fastvit}. 

Through extensive experimentation, our Lite-SAM algorithm achieves an optimal balance between model complexity and inference speed. 
In our research, we introduce Lite-SAM, a lightweight algorithm that capitalizes on the LiteViT backbone and leverages a prompt-based network architecture, 
namely AutoPPN. Lite-SAM distinguishes itself by having a low parameter count and reduced computational costs, 
yet it is capable of attaining performance benchmarks similar to those of SAM-B. 
Our comprehensive testing indicates that Lite-SAM strikes an optimal balance, 
offering reduced model complexity while maintaining swift inference speeds.

\section{Method: Lite-SAM}
\label{sec:litesam}

\begin{figure*}
	\centering
	\includegraphics[width=\linewidth]{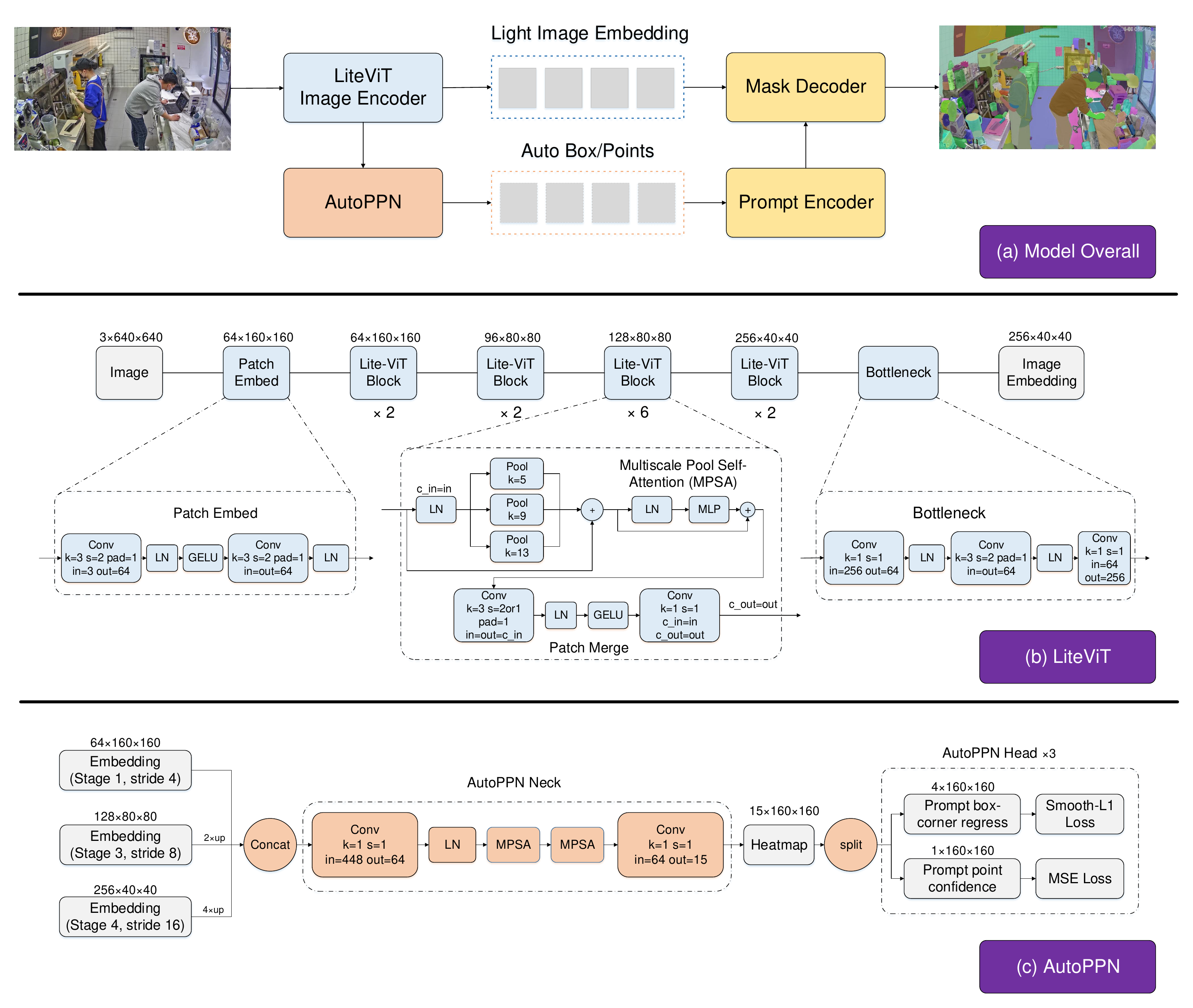}
    \caption{(a) Overview of the proposed Lite-SAM. The architecture consists of two detachable blocks,
			namely the Lightwight ViT backbone (LiteViT), Automated Prompt Proposal Network (AutoPPN).
		  (b) Macro Architecture of LiteViT. 	(c) Macro Architecture of AutoPPN. }
    \label{fig:litesam-overall}
\end{figure*}

\begin{figure*}
	\centering
	\begin{subfigure}{1\linewidth} 
			\includegraphics[width=1\linewidth]{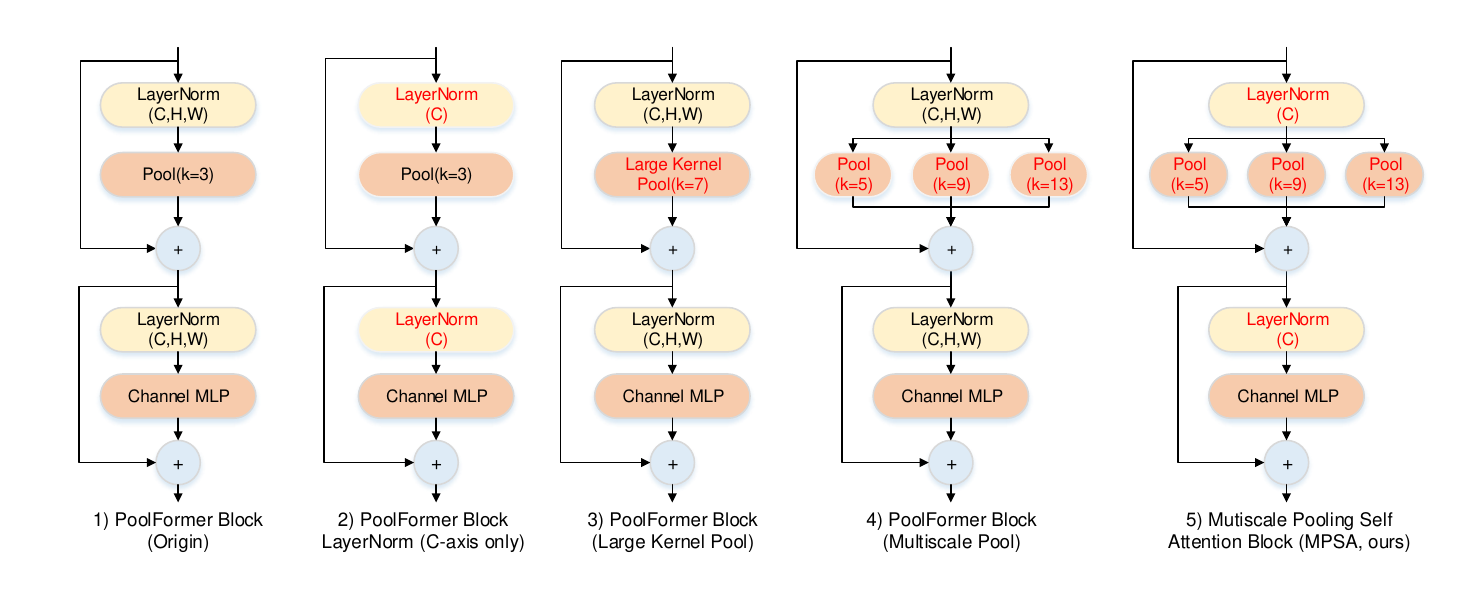}
			
		\end{subfigure}
		\caption{Overview of architectural choice. 
		(1) represents the original PoolFormer Block, 
		(2),(3) and (4) show the modifications to PoolFormer Block, 
		and (5) the final version of our Multiscale Pooling Self Attention module.} 
           \label{fig:blocks}
	\end{figure*}

\subsection{Design motivation and choices}

We present the Lite-SAM architecture, which consists of four main components:
a LiteViT encoder, an AutoPPN network, 
a standard prompt encoder, and a mask decoder as delineated in the SAM framework \cite{kirillov2023segment}. 
This configuration is visualized in \cref{fig:litesam-overall} (a). 
The novel AutoPPN module was specifically designed to streamline the automated prompt task. 
It simultaneously regresses both bounding box prompts and point prompts in an end-to-end fashion, 
which significantly cuts down the inference time for the SegEvery task 
when compared with dense positional encoding schemes from previous research. 
This advancement is key to achieving real-time segmentation. 
A comprehensive overview of the architecture and 
training methodologies will be provided in the forthcoming sections.

\subsection{LiteViT Architecture} \label{litevit}

Standard self-attention
token mixers  \cite{dosovitskiy2020image} are known for their high computational cost.
In contrast, the combined CNN-Transformer 
hybrid structure plays an essential role in crafting 
lightweight Vision Transformer (ViT) networks \cite{mehta2022mobilevit,liu2023efficientvit,vasu2023fastvit,yu2022metaformer}.
This hybrid balances model accuracy with computational efficiency. 
Inspired by efficient variations of self-attention layers in existing research, we have 
developed our LiteViT image encoder, beginning with a PoolFormer-S12 \cite{yu2022metaformer} baseline.  
We closely examine our architectural decisions, 
as detailed in \cref{tab:litevit_ablation_study} and illustrated in \cref{fig:blocks}.
As a supplement, we have also demonstrated the excellent scalability of LiteViT in \cref{tab:litevit_ablation_study}.

\begin{table}[t]
	\small\centering
	\caption{\textbf{LiteViT Attention block Ablation Studies.} All
models are trained and benchmarked using the same settings described in \cref{subsec:Hyperparameters}, with unified input resolution 640$\times$640.
As a supplementary addition, we have meticulously documented the performance metrics of LiteViT, specifically its floating-point operations (FLOPs), latency, and evaluation metrics, when scaled to 2 and 3 times the parameter volume of the baseline LiteViT network. Notably, this scaling achieves impressive mAP scores of 56.9\% and 58.1\% for 1-box prompt segmentation on the COCO dataset, respectively. These results underscore the commendable scalability of LiteViT.}
	\label{tab:litevit_ablation_study}
	\setlength{\tabcolsep}{4pt}{
		\scalebox{0.7}{
			\begin{tabular}{c | c | c | c | c | c | c}
				\toprule
				\makecell[c]{Architectual} & \makecell[c]{Attention\\ Block\\ Choices} & \makecell[c]{Params $\downarrow$\\(M)} & \makecell[c]{MACs $\downarrow$\\(G)} & \makecell[c]{Backbone \\ Latency $\downarrow$ \\ (ms/1-batch) } &  \makecell[c]{COCO 1-box\\ prompt mAP $\uparrow$\\(\%)}  &  \makecell[c]{Stages 1-4\\Embedding\_dims}\\
				\midrule
				\makecell[c]{PoolFormer-S12 \\ (Baseline~\cite{yu2022metaformer})}& \cref{fig:blocks} (1) & 11.9 & 45.2 & 30  & 55.1 &  \makecell[c]{[64, 128, 320, 512]} \\ %
				\midrule
				\multirow{5}{*}{\makecell[c]{ PoolFormer-S12-Tiny \\ (Embedding\_dims pruned)}}  & \makecell[c]{ \multirow{2}{*}{\cref{fig:blocks} (1)}  }  & 0.54 &  4.2 & 7.4  & 50.9 &  \makecell[c]{[32, 64, 96, 128]} \\ %
				\cmidrule{3-7}
				& & 1.15 & 10.8 & 8.4  & 52.7 &  \multirow{6}{*}{\makecell[c]{[64, 96, 128, 256]}} \\ %
				\cmidrule{2-6}				
				    &  \cref{fig:blocks} (2)           & 1.15 & 10.8 & 8.1  & 53.1 &    \\ %
				\cmidrule{2-6}	
				    &   \cref{fig:blocks} (3)           & 1.15 & 10.8 & 8.6  & 54.0 &    \\ %
				\cmidrule{2-6}	
				   &   \cref{fig:blocks} (4)            & 1.16 & 10.9 & 8.7  & 55.2 &    \\ %
				\cmidrule{1-6}	
				\makecell[c]{LiteViT\\(ours, \cref{litevit}) }  &  \cref{fig:blocks} (5)
                        & \makecell[c]{1.16 \\ \textbf{\textcolor{lightblue}{(-90$\%$)}}} 
                        & \makecell[c]{10.9  \\ \textbf{\textcolor{lightblue}{(-76$\%$)}}}
				  & \makecell[c]{8.6 \\ \textbf{\textcolor{lightblue}{(3.4x up $\uparrow$)}}} 
                       & \makecell[c]{55.3 \\ \textbf{\textcolor{lightblue}{(+0.2$\%$ $\uparrow$)}}} &     \\
				\cmidrule{1-7}	
				\makecell[c]{LiteViT \\ ($\sim$2$\times$ parameters) }  &  \cref{fig:blocks} (5)
                        & \makecell[c]{2.19  } 
                        & \makecell[c]{22.3  }
				  & \makecell[c]{12.4 } 
                       & \makecell[c]{56.9 } &  [96, 128, 192, 384]   \\
                     \cmidrule{1-7}	
				\makecell[c]{LiteViT \\ ($\sim$3$\times$ parameters) }  &  \cref{fig:blocks} (5)
                        & \makecell[c]{3.63 } 
                        & \makecell[c]{38.0 }
				  & \makecell[c]{15.9 }
                       & \makecell[c]{58.1 } &  [128, 160, 256, 512]   \\
				\bottomrule
			\end{tabular}
	}}
\end{table}

We base our image encoder model on a novel building block, referred to as the LiteViT Block. 
The detailed architectural specifications can be found in \cref{fig:litesam-overall} (b).
To overcome the challenge of capturing local features, we incorporate multiscale pooling into our lightweight 
attention module. Specifically, we introduce the Multi-Scale Pooling Module (MSPM) module to enhance the receptive field at each 
stage of the network architecture efficiently.

Within a LiteViT Block, the input is first processed by the MSPM module, followed by a convolutional 
MLP (Multilayer Perceptron) module; each stage is connected via skip connections. To facilitate downsampling and adjust the output channels at 
each stage, we employ a dedicated module known as the Patch Merge module, which effectively acts as a stem convolutional layer.

\subsection{AutoPPN}\label{sec:autoppn}

The standard approach of using dense positional encoding for prompts 
may not be suitable for real-time segmentation tasks due to the processing time required. 
To enhance the inference performance of the SegEvery task, 
we introduce the AutoPPN module, the architecture is detailed in \cref{fig:litesam-overall} (c).

It has been well-established that representing objects by a single point located at the center of their bounding box is a straightforward 
and efficient technique \cite{law2019cornernet,zhou2019objects}. 
Building on this concept, our AutoPPN framework predicts both prompt points and bounding boxes in an 
end-to-end manner from the output feature map. The corresponding loss is composed of two elements: confidence in the point prompt and 
accuracy in the bounding box regression. We have implemented three significant modifications to refine our approach, which are detailed below:

\begin{table}[t]
	\small\centering
	\caption{\textbf{AutoPPN Ablation Studies.} All
models are trained and benchmarked using the same settings described in \cref{subsec:Hyperparameters}.}
	\setlength{\tabcolsep}{4pt}{
		\scalebox{0.8}{
			\begin{tabular}{c | c | c | c| c }
				\toprule
				\makecell[c]{PPN Architectural Choices \\ \cref{sec:autoppn}} &   \makecell[c]{Stem Conv \\ $\to$ MSPM(\cref{mod1})} & \makecell[c]{ New GT \\ $\&$ Loss(\cref{mod2}) } & \makecell[c]{Object \\ Grouping(\cref{mod3})} & \makecell[c]{Mask \\ AR@1000($\%$)}  \\
				\midrule
                     \makecell[c]{Baseline = Stem Conv \\ + Focal-Loss/Smooth-L1 Loss \\ + w/o Object Grouping} & - & - & - & 48.8 \\
				\midrule
                     \makecell[c]{\multirow{3}{*}{1 improvement strategy}}  & \checkmark & - & - & 49.5 \\ 
                              & - & \checkmark & - & 50.1 \\
                              & - & - & \checkmark & 49.7 \\
                     \midrule 
                     \makecell[c]{\multirow{3}{*}{2 improvement strategies}}         & \checkmark & \checkmark & - & 51.4 \\
                              & \checkmark & - & \checkmark & 52.3\\
                              & - & \checkmark & \checkmark & 51.1 \\
				\midrule
                     \makecell[c]{AutoPPN \\(all improvement strategies)} & \checkmark & \checkmark & \checkmark & 53.0 \\                     
				\bottomrule
			\end{tabular}
	}}
	\label{tab:lite_sam_autoppn}
\end{table}

\begin{table}[t]
	\small\centering
	\caption{\textbf{Comparison of speed and accuracy acceleration of AutoPPN in SOTA models.}
		To ensure a fair comparison, we conducted AutoPPN training on both SAM and MobileSAM using the same data and training parameters.}
	\setlength{\tabcolsep}{4pt}{
		\scalebox{0.8}{
			\begin{tabular}{l | c | c | c  }
				\toprule
				Model  & Sampling Strategy  &  \makecell[c]{SegEvery Time$\downarrow$ \\(ms)} &  \makecell[c]{ COCO \\ AR@1000 $\uparrow$ ($\%$)} \\
				\midrule
				SAM-B    \cite{kirillov2023segment}     & Grid-Search (32 x 32)  & 2084 & 55.1  \\
				SAM-B + AutoPPN                         & AutoPPN(256 points)            & 120 \textbf{\textcolor{lightblue}{(17.3x up$\uparrow$)}} & 54.7 \\
				\midrule
				MobileSAM    \cite{zhang2023faster}     & Grid-Search (32 x 32)  & 2500 & 53.2 \\
				MobileSAM + AutoPPN                     & AutoPPN(256 points)            & 115 \textbf{\textcolor{lightblue}{(21.7x up$\uparrow$)}} & 52.6 \\
				\midrule
				\textbf{LiteViT(ours)}      & Grid-Search (32 x 32)  & 1320 & 53.4 \\
				\textbf{LiteViT + AutoPPN (ours,Lite-SAM)}  & AutoPPN(256 points)            & 80 \textbf{\textcolor{lightblue}{(16.5x up$\uparrow$)}}  & 52.8 \\
				\bottomrule
			\end{tabular}
	}}
	\label{tab:sam_autoppn} 
\end{table}

\begin{figure*}[!ht]
	\centering
	\begin{subfigure}{0.8\linewidth} 
			\includegraphics[width=1\linewidth]{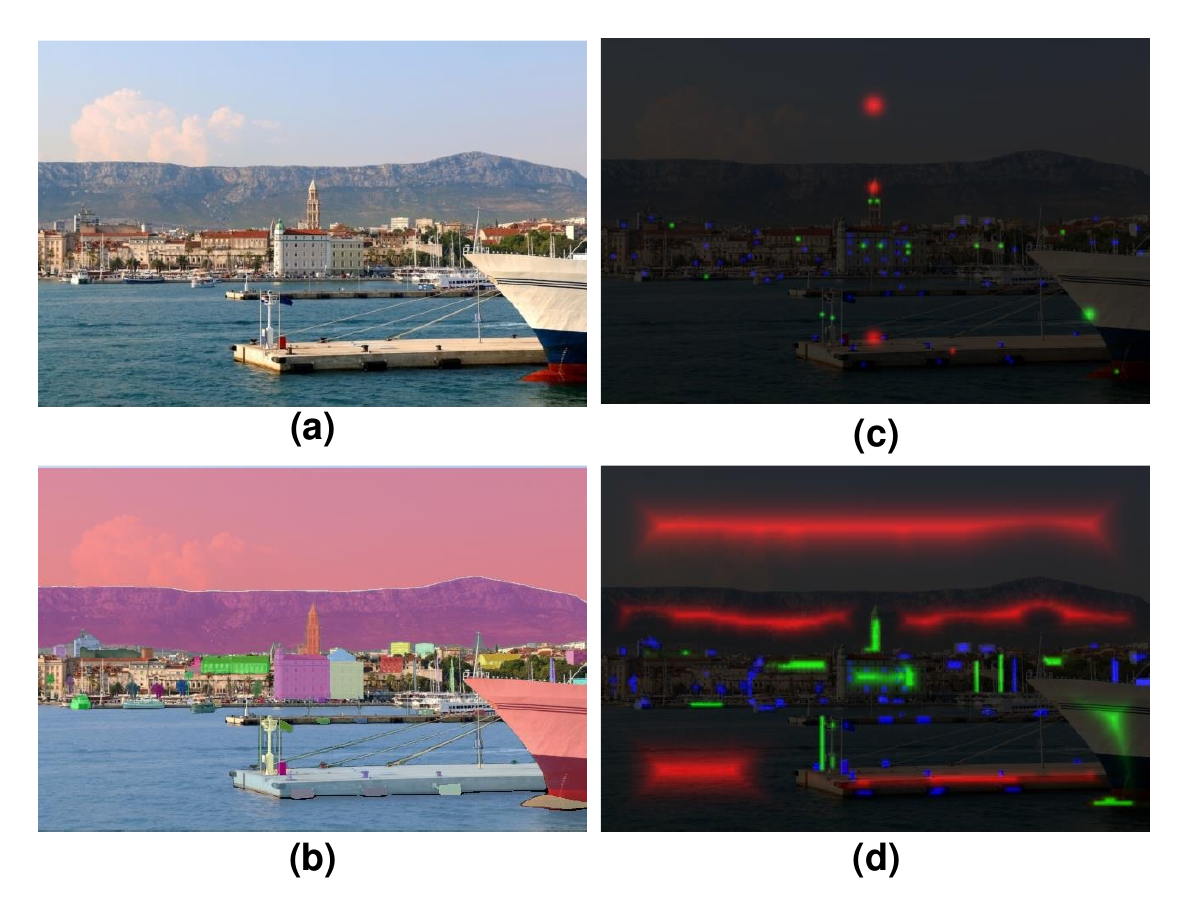}
		\end{subfigure}
		\caption{We compare two methods of generating pointwise foreground/background labels within an image (sa$\_$3196.jpg) from SA-1B \cite{kirillov2023segment} (a). 
		All the masks are visualized as shown in (b). 
The pointwise labels generated by large, medium, small masks, are visualized with red, green and blue color, respectively.
Comparing with bounding box center with gaussian kernel approach (c),  
distance transform approach (d) provides a more statisfactory result with less ambiguity.  
}
     \label{fig:hmapcompare}
\end{figure*}

\begin{enumerate}[(1)]

\item  \label{mod1} We have enhanced our network by replacing the basic stem convolution 
network with a more sophisticated stem MSPM network. 
This updated network effectively integrates multiscale spatial information,
 which significantly boosts the detection recall for large-scale objects or entities, 
 such as the sky, buildings, and water bodies.

\item  \label{mod2} To estimate the confidence of point prompts, we have incorporated the use of distance transforms. This facilitates the calculation of 
the distance between a point and its corresponding mask, as depicted in \cref{fig:hmapcompare}. 
In cases where a point falls within multiple masks, 
we select the one with the smallest area. The pseudo code for this procedure is provided in supplementary materials.
 
Unlike the Gaussian-based method 
referenced in \cite{zhou2019objects}, our technique enables the creation of a softened pointwise ground-truth distinction between foreground and background. 
Additionally, our method prioritizes the identification of the most central points of objects or entities rather than simply focusing on 
the center of their bounding boxes. This modification helps to alleviate the ambiguity present in scenarios involving unclear point prompts.
When computing loss, we have opted for a hard mining Mean Squared Error (MSE) Loss instead of the commonly used Focal-Loss for point prompt estimation.  

Meanwhile, the Smooth-L1 Loss remains the same as stated in \cite{law2019cornernet} for bounding box regression. 
It is also important to note that any unlabeled regions are excluded 
from the loss computation process. During inference, we only apply point-based non-maximum suppression (Point-NMS) and adhere to the practice 
of selecting the Top N points, as described in \cite{zhou2019objects}.

\item  \label{mod3} During the end-to-end regression stage, 
we divided the target masked regions into three groups based on the size of their bounding rectangles: 
large ($max(\frac{h}{H_{img}}, \frac{w}{W_{img}}) \geq 0.25$), 
medium ($max(\frac{h}{H_{img}}, \frac{w}{W_{img}}) \in (0.05, 0.25)$), 
and small ($max(\frac{h}{H_{img}}, \frac{w}{W_{img}}) \leq 0.05$). Separate loss calculations were performed for each group. 
The three improvements greatly enhance the performance, see \cref{tab:lite_sam_autoppn}. 
We denote 
\begin{equation} 
L_{ppn} = L_{H-MSE} + L_{S-L1}  \label{eq:ppn}
\end{equation}
the total loss of PPN regression, 
where $L_{H-MSE}$ refers to the hard mining MSE Loss and $L_{S-L1}$ the Smooth L1 Loss for box regression.
\end{enumerate}

\subsection{Total loss}

For the comprehensive training of Lite-SAM, we incorporate the mask loss,
which combines the original Focal-Loss \cite{lin2018focal} and Dice-Loss \cite{lin2018focal} from SAM \cite{kirillov2023segment}.
This combination quantifies the pixel-wise alignment between the predicted mask and the ground truth.
Additionally, a mean squared error loss measures the discrepancy between the IoU prediction and the intersection of the predicted mask with the ground truth mask.
The mask loss is formally expressed as:
\begin{equation}
	L_{mask} = \lambda_{f} L_{focal} + \lambda_{d} L_{dice} + \lambda_{i} L_{iou} \label{eq:mask}
\end{equation}
With \cref{eq:ppn} and \cref{eq:mask}, the total training loss is defined by
$L_{total} =  L_{ppn} + L_{mask}$.

\section{Experiments}
\label{sec:Experiments}

In this section, we present a comprehensive evaluation of our proposed Lite-SAM framework. To ensure a rigorous and equitable comparison, 
we utilized the same evaluation protocol as employed by other SOTA methods.

\subsection{Datasets}

\textbf{Public data.}
Lite-SAM was trained on SA-1B \cite{kirillov2023segment}. 
We selected three public datasets to assess the zero-shot capabilities of our model:
MSCOCO 2017 \cite{lin2014microsoft}, LVIS \cite{gupta2019lvis}, and BSDS500 \cite{martin2001database}.

\subsection{Implementation details}\label{subsec:Implementation}

\textbf{Hyperparameters.}\label{subsec:Hyperparameters}
We developed Lite-SAM using the PyTorch framework and trained it on 128 NVIDIA A40 GPUs, achieving an aggregate batch 
size of 256. The model underwent training from scratch, without the use of any pre-existing weights. 
With just 18$\%$ of the SA-1B dataset \cite{kirillov2023segment} dataset,
our model demonstrated impressive results. We utilized the Adam optimizer with an initial learning rate of 4e-5 and completed the training 
process in 4 epochs, which took a total of 50 hours. Throughout the training, all images were resized to 640$\times$640 pixels.
Concurrently, it is essential to recognize that the choice of using 18$\%$
of the SA-1B data was based on a trade-off between training
time and accuracy. The ablation study results regarding the selection of training data size and backbones, 
are presented in our supplementary material.

For supervising the guided prompt predictions, our loss function, AutoPPN-Loss, 
included a mix of hard mining MSE Loss for pointwise objectness
and L1-Loss for prompt box regression, with a respective ratio of 2:1.
For the mask prediction component, we employed a blended loss function combining Focal-Loss 
\cite{lin2018focal} and Dice-Loss \cite{lin2018focal} with a weighting of 10:1. 
In addition, a mean squared error loss was introduced to estimate the accuracy of the intersection over union (IoU) prediction compared to the 
ground truth mask alignment. The mean Intersection over Union (mIoU) metric was selected as our evaluation standard for segmentation performance.

\subsection{Comparison of speed and accuracy acceleration of AutoPPN in SOTA models}

As shown in \cref{tab:sam_autoppn}, the integration of AutoPPN leads to appreciable improvements in SegEvery time, while preserving the recall rates. 
Specifically, with the SAM-B \cite{kirillov2023segment} model, AutoPPN achieves a speedup of 17.3-fold relative to that of a conventional Grid Search method. 
For MobileSAM \cite{zhang2023faster}, the speedup stands at 21.7-fold. When applied to our LiteViT model, AutoPPN manages a speedup of 16.5-fold, 
reducing the SegEvery time to less than 80 ms, a significant milestone. These experimental results compellingly illustrate the efficiency of 
AutoPPN in addressing the speed bottleneck typically associated with Grid Search.

\begin{table}[t]
	\centering
	\caption{\textbf{Comparison with SOTA lightweight backbone models on COCO.}
		(1) All experiments are conducted based on open-source models and trained from scrach with same implementation described in \cref {subsec:Implementation}. 
		(2) The time tests for SegAny were conducted on an A40 GPU, while keeping the same environment.
		(3) In supplementary material, we present the classifcation capability of LiteViT on the ImageNet dataset to serve as a pre-trained model.}
	\label{tab:litevit_compare_modules}	
	\setlength{\tabcolsep}{4pt}{
		\scalebox{0.8}{
			\begin{tabular}{l | c |c |c | c | c | c |c | c }
				\toprule
				\multirow{2}{*}{Backbone Model}& \multicolumn{4}{c|}{COCO} & \multirow{2}{*}{SegAny time}   & \multirow{2}{*}{Params}  & \multirow{2}{*}{MACs} & \multirow{2}{*}{Input} \\
				\cmidrule{2-5}  & \makecell[c]{1-box\\(mAP)}   & \makecell[c]{1-box\\(mIou)}   & \makecell[c]{1-point\\(mAP)}  & \makecell[c]{1-point\\(mIou)}& (ms)  & (M)  &  (G)  &  \\
				\midrule
				Mobilenetv2~\cite{sandler2018mobilenetv2,howard2019searching}    & 48.2$\%$       & 69.6$\%$ & 23.5$\%$ & 48.5$\%$  & 5.1    & 1.89  & 4.0  & 640 \\
				Shufflenetv2~\cite{zhang2018shufflenet,ma2018shufflenet}          & 49.2$\%$       & 70.6$\%$ & 24.2$\%$ & 49.6$\%$  & 5.8    & 1.52  & 5.1  & 640 \\
				\midrule                                                             
				MobileViT~\cite{mehta2022mobilevit}                             & 51.4$\%$       & 72.1$\%$ & 26.2$\%$ & 53.8$\%$  & 11.9   & 5.57  & 13.7 & 640 \\
				EfficientViT~\cite{liu2023efficientvit}                           & 54.1$\%$       & 73.6$\%$ & 28.4$\%$ & 53.1$\%$  & 18.0   & 30.73 & 106.5& 640 \\
				FastViT~\cite{vasu2023fastvit}                                   & 52.3$\%$       & 70.0$\%$ & 28.0$\%$ & 48.0$\%$  & 7.5    & 3.98  & 7.0  & 640 \\              
				TinyViT~\cite{wu2022tinyvit}                                      & 54.0$\%$       & 73.4$\%$ & 26.5$\%$ & 52.6$\%$  & 17     & 6.07  & 36.6 & 640 \\
				\midrule                                                          
				\textbf{LiteViT(ours,backbone)}                                   & \textbf{55.8$\%$}       & \textbf{74.8$\%$}  & \textbf{32.9$\%$} & \textbf{55.3$\%$} & 7.9    & \textbf{1.16}  & 10.2  &  640 \\
				\bottomrule
			\end{tabular}
	}}
\end{table}

\subsection{Comparison with SOTA lightweight models on COCO 2017}

In \cref{tab:litevit_compare_modules}, we conducted detailed experimental comparisons and found that different backbone models exhibit varying levels of 
performance across each metric. Among these models, our proposed LiteViT (which serves as our backbone model) outperformed the other 
lightweight backbone models in all metrics and achieved the best results. Specifically, LiteViT reached a performance of 55.8$\%$ for 1-box mAP, 
74.8$\%$ for 1-box mIoU, 32.9$\%$ for 1-point mAP, and 55.3$\%$ for 1-point mIoU. Furthermore, LiteViT demonstrated clear advantages in terms of 
inference time, model parameter count, and computational load compared to other models. Overall, our experimental results establish that 
LiteViT, serving as our backbone model, is an exceptional lightweight backbone option, achieving SOTA performance on the COCO 
dataset. It also offers faster inference times and a relatively smaller model size. These results confirm its effectiveness and its competitive edge.

\begin{table*}[t]
	\small\centering
	\caption{\textbf{Zero-Shot Image Segmentation Results on MSCOCO 2017 and LVIS validation sets using mIoU and AP metric.}  
		(1) Note that the 1-box prompt result is not avaliable in Semantic-SAM-L's released code (similarly hereinafter). 
		(2)``r640'' means the input resolution is 640$\times$640. 
		(3) Note: We adopted the method in reference \cite{caiefficientvit}, which is entirely based on the ground truth (GT) box for predictions, instead of using ViTDet-H's results as prompts \cite{kirillov2023segment}. }
	
	\setlength{\tabcolsep}{4pt}{
		\scalebox{0.7}{
			\begin{tabular}{l | c | c | c | c | c c c c | c c c c }
				\toprule
				\multirow{2}{*}{Model} & \multicolumn{2}{c|}{MSCOCO(mIoU) $\uparrow$}& \multicolumn{2}{c|}{LVIS(mIou) $\uparrow$} & \multicolumn{4}{c|}{MSCOCO$\uparrow$} & \multicolumn{4}{c}{LVIS $\uparrow$}  \\
				\cmidrule{2-13}   & \makecell[c]{1-box \\ ($\%$)} & \makecell[c]{1-point\\($\%$)} & \makecell[c]{1-box \\($\%$)} & \makecell[c]{1-point\\($\%$)} &\makecell[c]{AP\\($\%$)} & \makecell[c]{AP\textsuperscript{S}\\($\%$)} & \makecell[c]{AP\textsuperscript{M}\\($\%$)} & \makecell[c]{AP\textsuperscript{L}\\($\%$)} & \makecell[c]{AP\\($\%$)} & \makecell[c]{AP\textsuperscript{S}\\($\%$) }& \makecell[c]{AP\textsuperscript{M}\\($\%$)} & \makecell[c]{AP\textsuperscript{L}\\($\%$)}\\
				\midrule
				SAM-B    \cite{kirillov2023segment} & 75.0    & 52.2 & 73.5 & 55.2 & 56.6 & 47.4 & 60.3 & 68.0 & 61.1 & 50.3 &71.6 & 76.7\\
				SAM-L    \cite{kirillov2023segment} & 76.4    & 56.8 & 75.0 & 55.8 & 59.4 & 48.8 & 65.3 & 72.1 & 64.9 & 53.5 &76.1 & 81.9\\
				SAM-H    \cite{kirillov2023segment} & \textbf{76.5}    & \textbf{57.4} & \textbf{75.3} & \textbf{56.4} & \textbf{59.8} & 49.4 & 63.8 & 71.9 & \textbf{65.2} & 53.6& 76.5 & 82.1\\
				Semantic-SAM-L \cite{li2023semantic}  & N/A & 54.7 & N/A & 34.8     & \multicolumn{8}{c}{N/A} \\
				Fast-SAM   \cite{zhao2023fast}              & 72.8 & 50.2 & 67.3 & 46.8 & 47.5 & 37.9 & 48.1 & 56.4  & 43.8 & 35.1 & 45.6 & 59.7\\
				EfficientViT-L0-SAM \cite{caiefficientvit}  & 74.5 & 51.3 & 73.1 & 52.9 & 56.1 & 44.3 & 59.7 & 70.8  & 59.8 & 46.8 & 70.2 & 80.1\\
				EfficientViT-L1-SAM \cite{caiefficientvit}  & 75.2 & 51.5 & 73.9 & 54.8 & 57.1 & 45.4 & 60.8 & 71.5  & 61.4 & 48.0 & 72.5 & 81.6\\
				Mobile-SAM-v2* \cite{zhang2023mobilesamv2}  & 72.8 & 50.5 & 67.7 & 42.4 & 51.4  & 41.6 & 55.1 & 64.1 & 52.8 & 42.2 & 63.2 & 69.6\\
				\midrule
				Mobile-SAM \cite{zhang2023faster}           & 72.8 & 50.5 & 67.7 & 42.4 & 51.4  & 41.6 & 55.1 & 64.1 & 52.8 & 42.2 & 63.2 & 69.6\\
				Edge-SAM \cite{zhou2023edgesam}             & 74.0 & 51.9 & 69.4 & 43.8 & 52.5  & 42.7 & 56.0 & 65.3 & 54.1 & 43.5 & 63.9 & 70.7\\
				\textbf{Lite-SAM(ours, r640)}               & 74.8 & 55.8 & 73.2 & 54.4 & 55.8  & 46.7 & 59.6 & 69.6 & 58.4 & 45.9 & 66.9 & 77.5\\
				\textbf{Lite-SAM(ours, r1024)} & \textbf{76.3} &\textbf{56.9} & \textbf{75.7} & \textbf{57.3} & \textbf{56.5} & 47.4  & 61.0 & 70.7 &  \textbf{60.7} & 49.3 & 71.9 &  79.8\\
				\bottomrule
			\end{tabular}
	}}
	\label{tab:sam_coco_livs_miou}
\end{table*}

\subsection{Comparison with SOTA Algorithms on COCO and LVIS validation sets using AP and mIou metric}
In \cref{tab:sam_coco_livs_miou},  we conducted detailed experimental comparisons among the latest algorithms from the SAM series on the COCO and LVIS datasets. 
The results show that the SAM-H \cite{kirillov2023segment}  model achieved superior performance, obtaining the highest metrics on both datasets. Specifically, 
its 1-box mIoU reached 76.5$\%$ on the COCO dataset and 75.3$\%$ on the LVIS dataset. In contrast, the 1-point mIoU scored 57.4$\%$ for COCO and 56.4$\%$ for LVIS.

The experimental results classify the models into two categories based on their size: large models with a parameter count exceeding 10M and 
lightweight models with fewer than 10M parameters. Among the large models are SAM-B/L/H \cite{kirillov2023segment}, Semantic-SAM-L \cite{li2023semantic}, Fast-SAM \cite{zhao2023fast}, EfficientViT-L0-SAM \cite{caiefficientvit}, 
MobileSAM-v2 \cite{zhang2023mobilesamv2}, and EfficientViT-L1-SAM \cite{caiefficientvit}. The lightweight category includes Mobile-SAM \cite{zhang2023faster}, Edge-SAM \cite{zhou2023edgesam} , and Lite-SAM.

Lite-SAM, a lightweight model, achieved a 1-box mIoU performance that surpassed SAM-B \cite{kirillov2023segment} by 1.3$\%$, with significantly fewer parameters and computational 
demands. Lite-SAM also outperformed Mobile-SAM \cite{zhang2023faster} and Edge-SAM \cite{zhou2023edgesam}  in terms of mIoU metrics. Regarding the Average Precision (AP) metric, SAM-H \cite{kirillov2023segment} still 
recorded the highest values, with an AP of 59.8$\%$ on COCO and 65.2$\%$ on LVIS. Lite-SAM performed better than Mobile-SAM \cite{zhang2023faster} and Edge-SAM \cite{zhou2023edgesam}   but slightly 
lower than EfficientViT-L1-SAM \cite{caiefficientvit} in terms of AP.

Overall, these experiments highlight the outstanding performance of our Lite-SAM algorithm, confirming its effectiveness and competitive edge in the field.

\begin{table}[t]
	\small\centering
	\caption{\textbf{Comparison with SOTA Algorithms: Model Complexity, SegEvery Speed, and Mask AR@1000 metric Evaluation on COCO2017.}
		(1) ``r640'' means the input resolution is $640 \times 640$. For specific calculation details, please refer to the code examples in the supplementary materials.
		(2) * The ``Fast-SAM'' does not have true interactive segmentation via point or box prompts. It employs heuristic rules for post-process object selection, a
method that aligns marginally with the SAM principles. Therefore, the Fast-SAM algorithm and the other SAM series algorithms are completely different, making a comparison between them of no value.
		(3) ** The ``Mobile-SAM-v2''  paper does not include statistics on the parameter size and computational complexity of the Object-aware model, so the data has been re-estimated (43M = Yolov8(33.3M) + MobileSAM(9.7M)).
     }
	\label{tab:sam_complex_segevery}
	\setlength{\tabcolsep}{4pt}{
		\scalebox{0.7}{
			\begin{tabular}{l | c | c | c | c | c | c | c}
				\toprule
				Model& Params $\downarrow$ & MACs $\downarrow$ & \makecell[c]{SegEvery \\ Time $\downarrow$} & \makecell[c]{Mask AR\\@1000 (\%)} & Sampling Strategy & Train Strategy & Year\\
				\midrule
				SAM-B    \cite{kirillov2023segment}        & 90M  & 371G & 2.1s  & 55.1 & Grid-Search (32 x 32) & pretrain on MAE & 2023\\
				SAM-L    \cite{kirillov2023segment}        & 308M & 1.3T & 3.3s  & 56.6 & Grid-Search (32 x 32) & pretrain on MAE & 2023\\
				SAM-H    \cite{kirillov2023segment}        & 635M & 2.7T & 3.5s  & 58.7 & Grid-Search (32 x 32) & pretrain on MAE & 2023\\
				Semantic-SAM \cite{li2023semantic}         & 202M & 1.4T & 2.6s  & 55.0 & Grid-Search (32 x 32) & from scratch & 2023\\				
				EfficientViT-L0-SAM \cite{caiefficientvit} & 31M  & 109G & 1.7s  & 56.7 & Grid-Search (32 x 32) & from scratch & 2023\\
                     \midrule
                     Fast-SAM \cite{zhao2023fast} *        & 72.2M & 443G & 0.04s  & 53.3 & \makecell[c]{Post-Process \\Object Selection} & \makecell[c]{pretrain on \\YOLOv8} & 2023\\
				\midrule
				Mobile-SAM \cite{zhang2023faster}          & 9.7M & 39.6G   & 2.5s  & 53.2 & Grid-Search (32 x 32)   & distillation  & 2023\\
				Edge-SAM \cite{zhou2023edgesam}            & 9.7M & 23.4G   & 1.6s  & 51.9 & Grid-Search (32 x 32)   & distillation & 2024\\
				\midrule
				Mobile-SAM-v2 \cite{zhang2023mobilesamv2} ** & \makecell[c]{43M}& 470G    & 0.13s   & 53.6 & Object-Aware  & distillation & 2024\\
				\midrule
				\textbf{Lite-SAM(r640)}                   & \textbf{4.2M}    & \textbf{12.7G}  &\textbf{0.08s} & 52.8 & AutoPPN(256 points) & from scratch & 2024\\
				\textbf{Lite-SAM(r1024)}                  & \textbf{4.2M}    & 32.5G           & 0.1s          & 54.1 & AutoPPN(256 points) & from scratch & 2024\\
				\bottomrule
			\end{tabular}
	}}
\end{table}

\subsection{Comparison with SOTA Algorithms Complexity and SegEvery Speed Evaluation}
In \cref{tab:sam_complex_segevery}, we have presented detailed experimental comparisons of the latest algorithms in the SAM series. The results reveal substantial 
variations in parameter size, Multiply-Accumulate Operations (MACs), and SegEvery runtime across the different algorithmic models. 
The Sampling Strategy is categorized into three types: Grid-Search, Object-aware, and AutoPPN.

SAM-B  \cite{kirillov2023segment}  boasts a parameter size of 90M, MACs of 371G, and a SegEvery runtime of 2.1s. The lightweight models, namely Mobile-SAM \cite{zhang2023faster} and Edge-SAM \cite{zhou2023edgesam}, 
have parameter sizes of 9.7M and offer 39.6/23.4G MACs, respectively. Mobile-SAM-v2 \cite{zhang2023mobilesamv2} implements the Object-aware strategy, leveraging YOLOv8 \cite{yolov8_ultralytics1} to 
perform box and point detection in advance, characterizing it as a two-stage algorithm.

Our newly developed Lite-SAM is designed as an end-to-end algorithm with a minimal parameter size of only 4.2M. Impressively, it has reduced the 
SegEvery runtime to a mere 80ms for the first time. This model not only demonstrates the best performance in regards to parameter size and MACs, 
but also in SegEvery inference time, which underlines its efficiency and competitive edge.

To demonstrate that Lite-SAM delivers results on par with other SAM architectures, 
while also showcasing its exceptional performance relative to other lightweight SAMs, 
we have included visual qualitative assessments for the ``SegEvery'' and ``SegAny'' tasks as supplementary material. 
These illustrations underscore the effectiveness of the Lite-SAM approach.

\begin{table}[!ht]
	\small\centering
	\caption{\textbf{Zero-shot transfer to edge detection on BSDS500.} Evaluation data of other methods is from \cite{kirillov2023segment}.}
     \label{tab:sam_bsds500}
	\setlength{\tabcolsep}{4pt}{
       \scalebox{0.9}
		{
			\begin{tabular}{lc| c c c c }
				\toprule
				Method & Year & ODS & OIS & AP & R50  \\
				\hline
				HED \cite{xie2015holistically} & 2015& 0.788 & 0.808 & 0.840 & 0.923 \\
				EDETR \cite{pu2022edter} & 2022 & 0.840 & 0.858 & 0.896 & 0.930 \\
				\multicolumn{6}{@{}l}{\emph{zero-shot transfer methods:}} \\
				Sobel filter & 1968 & 0.539 & - & -& - \\
				Canny \cite{canny1986computational} & 1986 & 0.600 &0.640 &0.580 & - \\
				Felz-Hutt \cite{felzenszwalb2004efficient} & 2004 &0.610 &0.640 &0.560 & - \\
				SAM-H \cite{kirillov2023segment}             & 2023 &0.768 &0.786 &0.794 & 0.928 \\
				Fast-SAM  \cite{zhao2023fast}       & 2023 &0.750 &0.790 &0.793 & 0.903 \\
				\midrule
				\textbf{Lite-SAM(ours)}                    & 2023 &0.761 &0.788 &0.793 & 0.919 \\
				\bottomrule
			\end{tabular}
	}}	
\end{table}

\subsection{Zero-Shot Edge Detection}
We assessed the zero-shot edge detection capability of Lite-SAM on the BSDS500 dataset \cite{martin2001database,arbelaez2010contour},
following the experimental parameters established 
by SAM \cite{kirillov2023segment} and Fast-SAM \cite{zhao2023fast}.
As shown in \cref{tab:sam_bsds500}, Lite-SAM R50 attains a metric score of 0.919, 
slightly behind SAM's 0.928 and surpassing Fast-SAM's 0.903.

\section{Conclusion}
\label{sec:conclusion}

In this paper, we propose an end-to-end lightweight algorithm called Lite-SAM,
which aims to address the high computational complexity issue of the SegEvery model in the SAM series.
Lite-SAM consists of the LiteViT module and the AutoPPN module, enabling modular deployment.
Our algorithm achieves a 16-fold speedup in inference time while maintaining a minimal decrease in accuracy compared to the SegEvery mode.
Through extensive experimental tests, we demonstrate that our approach satisfies the requirements of efficient and resource-friendly segmentation algorithms,
providing possibilities for practical applications in various fields.

\section*{Acknowledgements}

This work is supported by Zhejiang Dahua Technology Co., Ltd. and Zhejiang University. 

%
%
\bibliographystyle{splncs04}
\bibliography{main}

\newpage
\appendix

\section{More quantitative and qualitative evaluation and results}
\subsection{Image Classifcation: serving as a pre-trained model}

To understand the effectiveness of LiteViT backbone in image classifcation, we train our models on ImageNet following the
standard training strategy. We summarize the results and compare our models with SOTA image classifcation models in \cref{tab:imagenet}.
We have demonstrated that our LiteViT achieves an optimal balance of performance and accuracy in classification, establishing a new state-of-the-art (SOTA) standard.
LiteViT achieves an optimal balance between performance and accuracy. With only 1.16M parameters and 1.2G computational cost, it rivals the accuracy of models with over 20M parameters, 
even at a size of 224. This remarkable feat demonstrates the efficiency and effectiveness of LiteViT.

\begin{table*}[!ht]
	\small\centering
	\caption{\textbf{LiteViT Performance on ImageNet Classification.} 
		1) All these models are only trained on the ImageNet-1K training set and the accuracy on the validation set is reported. RSB-ResNet means the results are from ``ResNet Strikes Back''. }
	\label{tab:imagenet}
     \scalebox{0.9}
		{
	\begin{tabular}{l | c | c c | c c | c }
		\toprule
		\multirow{2}{*}{Models}  & \multirow{2}{*}{Top1 Acc(\%) $\uparrow$} & \multirow{2}{*}{Params(M) $\downarrow$} & \multirow{2}{*}{MACs(G) $\downarrow$}  & \multirow{2}{*}{Input Size}\\
		& & & &  \\
		\midrule
		RSB-ResNet-18 \cite{resnet,resnet_improved}                      & 70.6  & 12 & 1.8 &  224 \\
		RSB-ResNet-34 \cite{resnet,resnet_improved}                     & 75.5  & 22 & 3.7 &  224 \\
		RSB-ResNet-50 \cite{resnet,resnet_improved}                     & 79.8  & 26 & 4.1 &  224 \\
		\midrule
		MoblieViT-S \cite{mehta2021mobilevit}                      & 78.4  & 6  & 1.5 &  256 \\
		TinyViT-5M \cite{wu2022tinyvit}                        & 79.1  & 5.4 & 1.3 &  224 \\
		GLiT-Tiny \cite{glit}                         & 76.3  & 7   & 1.5 &  224 \\
		ViTAS-DeiT-A \cite{vitas}                      & 75.5  & 6   & 1.3 &  224 \\
		PoolFormer-S12 \cite{yu2022metaformer}                    & 77.2  & 12 & 1.8 &  224 \\
		EfficientViT \cite{caiefficientvit}                       & 82.7  & 24 & 2.1 &  256 \\
		CoAtNet-0 \cite{dai2021coatnet}    & 81.6  & 25 & 4.2 &  224 \\
		ConvNeXt-T \cite{liu2022convnet}   & 82.1  & 29 & 4.5 &  224 \\
		DeiT-S \cite{touvron2021training}  & 79.8  & 22 & 4.6  &  224  \\
		PVT-Tiny \cite{pvt}                & 75.1  & 13 & 1.9  &  224  \\
		PVT-Small \cite{pvt}               & 79.8  & 25 & 3.8  &  224  \\
		ResMLP-S12 \cite{resmlp}           & 76.6  & 15 & 3.0  &  224  \\
		Swin-Mixer-T/D24 \cite{swin}       & 79.4  & 20 & 4.0  &  256 \\
		gMLP-S \cite{gmlp}                 & 79.6  & 20 & 4.5  &  224  \\
		ViT-L/16$^*$ \cite{dosovitskiy2020image}            & 76.1  & 307 & 63.6 &  224\\	
		\midrule
		\textbf{LiteViT(ours)}             & 78.5  & \textbf{1.16} & \textbf{1.2} &  224 \\
		\bottomrule
	\end{tabular}}
\end{table*}

\subsection{Class-wise comparative analysis of Lite-SAM with other SAM models}

\newcolumntype{C}[1]{>{\centering\arraybackslash}m{#1}} 
As a supplement to Section 4.5 (Table 5), we have compared the performance of our approach on COCO across 80 object classes with other three SAM architectures in \cref{tab:classwise_compare}. This comparison showcases the competitive results of Lite-SAM in relation to other SAM models.

\small
\begin{longtable}
{c  C{2cm} C{2cm} C{2cm} C{2cm} C{2cm} }
\caption{\textbf{Lite-SAM has achieved competitive results in both overall and class-wise performance.} The best results in each class are displayed in red.} \\
\label{tab:classwise_compare} \\
\toprule
\multirow{4}{*}{Category} & \multicolumn{5}{c}{Model} \\
\cmidrule(lr){2-6}
 & \makecell[c] {SAM-B\cite{kirillov2023segment}\\ (r1024)} & \makecell[c] {EfficientViT \\ L0-SAM\cite{caiefficientvit} \\(r1024)} & \makecell[c] {MobileSAM\cite{zhang2023faster} \\(r1024)} & \makecell[c] {Lite-SAM \\(r640)} & \makecell[c] {Lite-SAM \\(r1024)} \\
\midrule
\endfirsthead
\multicolumn{6}{c}%
{{\bfseries \tablename\ \thetable{} -- continued from previous page}} \\
\toprule
Category & SAM-B\cite{kirillov2023segment} (r1024) & EfficientViT-L0-SAM\cite{caiefficientvit} (r1024) & MobileSAM\cite{zhang2023faster} (r1024) & Lite-SAM (r640) & Lite-SAM (r1024) \\
\midrule
\endhead

\midrule
\multicolumn{6}{r}{{Continued on next page}} \\
\midrule
\endfoot

\bottomrule
\endlastfoot

 overall          &  \textcolor{red}{0.566}   &  0.561   &  0.540   &  0.558   &  0.565  \\
\midrule
 person           &  \textcolor{red}{0.544}   &  0.532   &  0.498   &  0.498   &  0.522  \\
 bicycle          &  0.294   &  0.276   &  0.247   &  0.283   &  \textcolor{red}{0.295}  \\
 car              &  0.576   &  0.539   &  0.511   &  0.567   &  \textcolor{red}{0.588}  \\
 motorcycle       &  0.420   &  \textcolor{red}{0.438}   &  0.365   &  0.364   &  0.390  \\
 airplane         &  0.570   &  \textcolor{red}{0.612}   &  0.576   &  0.525   &  0.543  \\
 bus              &  \textcolor{red}{0.779}   &  0.767   &  0.758   &  0.765   &  0.748  \\
 train            &  0.717   &  \textcolor{red}{0.739}   &  0.725   &  0.722   &  0.717  \\
 truck            &  \textcolor{red}{0.684}   &  0.660   &  0.638   &  0.659   &  0.661  \\
 boat             &  0.456   &  0.444   &  0.420   &  0.474   &  \textcolor{red}{0.504}  \\
 traffic light    &  0.527   &  0.498   &  0.509   &  \textcolor{red}{0.595}   &  0.592  \\
 fire hydrant     &  0.716   &  0.717   &  0.709   &  \textcolor{red}{0.720}   &  0.703  \\
 stop sign        &  0.790   &  0.777   &  0.753   &  \textcolor{red}{0.807}   &  0.766  \\
 parking meter    &  0.743   &  0.728   &  0.741   &  \textcolor{red}{0.784}   &  0.758  \\
 bench            &  \textcolor{red}{0.402}   &  0.398   &  0.371   &  0.367   &  0.398  \\
 bird             &  \textcolor{red}{0.434}   &  0.419   &  0.391   &  0.340   &  0.416  \\
 cat              &  0.683   &  \textcolor{red}{0.770}   &  0.744   &  0.673   &  0.675  \\
 dog              &  0.710   &  \textcolor{red}{0.745}   &  0.715   &  0.667   &  0.676  \\
 horse            &  \textcolor{red}{0.483}   &  0.478   &  0.440   &  0.421   &  0.438  \\
 sheep            &  0.551   &  \textcolor{red}{0.571}   &  0.511   &  0.533   &  0.534  \\
 cow              &  \textcolor{red}{0.587}   &  0.587   &  0.534   &  0.540   &  0.564  \\
 elephant         &  0.649   &  \textcolor{red}{0.680}   &  0.637   &  0.611   &  0.608  \\
 bear             &  0.748   &  \textcolor{red}{0.784}   &  0.777   &  0.764   &  0.744  \\
 zebra            &  0.575   &  \textcolor{red}{0.606}   &  0.562   &  0.525   &  0.535  \\
 giraffe          &  0.549   &  \textcolor{red}{0.571}   &  0.537   &  0.454   &  0.493  \\
 backpack         &  0.502   &  0.487   &  0.448   &  0.503   &  \textcolor{red}{0.513}  \\
 umbrella         &  \textcolor{red}{0.633}   &  0.630   &  0.605   &  0.586   &  0.612  \\
 handbag          &  \textcolor{red}{0.457}   &  0.443   &  0.423   &  0.437   &  0.450  \\
 tie              &  \textcolor{red}{0.484}   &  0.439   &  0.413   &  0.437   &  0.482  \\
 suitcase         &  0.648   &  0.675   &  0.655   &  \textcolor{red}{0.680}   &  0.676  \\
 frisbee          &  0.708   &  0.710   &  0.691   &  \textcolor{red}{0.728}   &  0.708  \\
 skis             &  \textcolor{red}{0.068}   &  0.062   &  0.051   &  0.029   &  0.065  \\
 snowboard        &  0.328   &  0.315   &  0.311   &  0.320   &  \textcolor{red}{0.386}  \\
 sports ball      &  0.612   &  0.590   &  0.580   &  0.628   &  \textcolor{red}{0.647}  \\
 kite             &  \textcolor{red}{0.506}   &  0.476   &  0.470   &  0.421   &  0.487  \\
 baseball bat     &  \textcolor{red}{0.436}   &  0.400   &  0.371   &  0.339   &  0.387  \\
 baseball glove   &  0.619   &  0.619   &  0.612   &  \textcolor{red}{0.650}   &  0.637  \\
 skateboard       &  \textcolor{red}{0.338}   &  0.331   &  0.317   &  0.313   &  0.337  \\
 surfboard        &  0.472   &  0.450   &  0.422   &  0.460   &  \textcolor{red}{0.493}  \\
 tennis racket    &  \textcolor{red}{0.562}   &  0.559   &  0.532   &  0.518   &  0.516  \\
 bottle           &  0.633   &  0.605   &  0.582   &  0.637   &  \textcolor{red}{0.641}  \\
 wine glass       &  0.452   &  0.437   &  0.405   &  0.430   &  \textcolor{red}{0.461}  \\
 cup              &  0.698   &  0.675   &  0.669   &  \textcolor{red}{0.721}   &  0.710  \\
 fork             &  0.262   &  \textcolor{red}{0.282}   &  0.201   &  0.190   &  0.240  \\
 knife            &  0.339   &  0.311   &  0.266   &  0.296   &  \textcolor{red}{0.356}  \\
 spoon            &  0.374   &  0.334   &  0.302   &  0.321   &  \textcolor{red}{0.374}  \\
 bowl             &  0.618   &  0.505   &  0.561   &  \textcolor{red}{0.664}   &  0.629  \\
 banana           &  0.597   &  0.567   &  0.581   &  0.600   & \textcolor{red}{0.604}   \\
 apple            &  0.653   &  0.645   &  0.637   &  \textcolor{red}{0.671}   &  0.656  \\
 sandwich         &  0.739   &  0.702   &  0.718   &  \textcolor{red}{0.746}   &  0.725  \\
 orange           &  0.670   &  0.653   &  0.654   &  \textcolor{red}{0.697}   &  0.680  \\
 broccoli         &  0.483   &  0.463   &  0.465   &  \textcolor{red}{0.534}   &  0.496  \\
 carrot           &  0.555   &  0.543   &  0.512   &  0.544   &  \textcolor{red}{0.563}  \\
 hot dog          &  0.561   &  0.578   &  0.555   &  \textcolor{red}{0.610}   &  0.594  \\
 pizza            &  0.666   &  \textcolor{red}{0.667}   &  0.655   &  0.666   &  0.652  \\
 donut            &  0.724   &  0.719   &  0.702   &  \textcolor{red}{0.746}   &  0.728  \\
 cake             &  0.685   &  0.672   &  0.684   &  \textcolor{red}{0.708}   &  0.680  \\
 chair            &  0.421   &  0.433   &  0.413   &  0.425   &  \textcolor{red}{0.433}  \\
 couch            &  0.562   &  0.583   &  0.555   &  0.584   &  \textcolor{red}{0.589}  \\
 potted plant     &  0.429   &  0.455   &  0.447   &  \textcolor{red}{0.467}   &  0.465  \\
 bed              &  0.465   &  0.415   &  0.461   &  \textcolor{red}{0.520}   &  0.479  \\
 dining table     &  0.220   &  0.214   &  0.228   &  0.274   &  \textcolor{red}{0.279}  \\
 toilet           &  0.689   &  \textcolor{red}{0.711}   &  0.700   &  0.690   &  0.684  \\
 tv               &  0.753   &  0.757   &  0.739   &  \textcolor{red}{0.763}   &  0.738  \\
 laptop           &  0.674   &  \textcolor{red}{0.716}   &  0.671   &  0.675   &  0.675  \\
 mouse            &  0.707   &  0.679   &  0.701   &  \textcolor{red}{0.721}   &  0.711  \\
 remote           &  0.528   &  0.489   &  0.470   &  0.487   &  \textcolor{red}{0.532}  \\
 keyboard         &  0.691   &  0.694   &  0.691   &  \textcolor{red}{0.715}   &  0.703  \\
 cell phone       &  \textcolor{red}{0.597}   &  0.557   &  0.531   &  0.559   &  0.585  \\
 microwave        &  0.776   &  \textcolor{red}{0.801}   &  0.771   &  0.797   &  0.762  \\
 oven             &  0.615   &  0.567   &  0.582   &  0.628   &  \textcolor{red}{0.632}  \\
 toaster          &  0.825   &  0.823   &  0.738   &  \textcolor{red}{0.836}   &  0.815  \\
 sink             &  0.611   &  0.603   &  0.578   &  \textcolor{red}{0.648}   &  0.638  \\
 refrigerator     &  \textcolor{red}{0.786}   &  0.776   &  0.772   &  0.786   &  0.733  \\
 book             &  0.416   &  0.410   &  0.361   &  0.452   &  \textcolor{red}{0.476}  \\
 clock            &  0.745   &  0.732   &  0.715   &  \textcolor{red}{0.803}   &  0.774  \\
 vase             &  0.685   &  0.653   &  0.657   &  \textcolor{red}{0.687}   &  0.672  \\
 scissors         &  \textcolor{red}{0.324}   &  0.305   &  0.247   &  0.284   &  0.277  \\
 teddy bear       &  0.659   &  \textcolor{red}{0.693}   &  0.667   &  0.633   &  0.647  \\
 hair drier       &  0.473   &  \textcolor{red}{0.600}   &  0.477   &  0.418   &  0.455  \\
 toothbrush       &  \textcolor{red}{0.369}   &  0.328   &  0.307   &  0.299   &  0.359  \\
\bottomrule
\end{longtable}

\subsection{Ablation study on the selection of training data}

In Section 4.2, our choice of using 18$\%$
of the SA-1B data was based on a trade-off between training
time and accuracy. The ablation study results regarding the selection of training data size and backbones, are presented in \cref{tab:datasize}.

\begin{table}[!ht]
	\small\centering
	\caption{\textbf{Ablation study on the selection of training data size.} See Section 4.2 for Implementation details. The evaluation metrics is 1-box prompt mAP on COCO2017(val) dataset. We finally chose $18\%$ of SA-1B as training data, exemplying an optimal balance between training time and accuracy. It should be noted that the results for SAM-B, EfficientViT-L0-SAM, and MobileSAM are all reproduced by our own, without any open-source SAM training code avaliable. Therefore there may be minor inconsistencies with the original papers or models.
 }
     \label{tab:datasize}
	\setlength{\tabcolsep}{4pt}{
      \scalebox{0.9}
		{
			\begin{tabular}{c | c | c | c | c | c }
			    \toprule
		          \multirow{3}{*}{Model} &  \multirow{4}{*}{ \makecell[c]{Metric \&  \\ Training time\\(4 epochs)} } & \multicolumn{4}{c}{Training images of SA-1B}  \\
		          \cmidrule{3-6}         &       & \makecell[c]{1M\\(9\%)} & \makecell[c]{2M\\(18\%)} &  \makecell[c]{5M\\(45\%)} & \makecell[c]{11M\\(100\%)}   \\
                     \midrule
				\multirow{2}{*}{\makecell[c]{SAM-B\cite{kirillov2023segment}\\(r1024)}}      &   mAP(\%)               & 51.9  &  54.4  &  57.4  & 59.0  \\
                                &   Hours       & 47    & 96   & 230  & 513 \\
                     \midrule
				\multirow{2}{*}{\makecell[c]{EfficientViT-L0-SAM\cite{caiefficientvit}\\(r1024)}}     &   mAP(\%)     & 52.1   & 55.6    & 56.7   & 57.9 \\
                                &   Hours       & 40    &  83  & 203  & 427 \\
                     \midrule
				\multirow{2}{*}{\makecell[c]{MobileSAM\cite{zhang2023faster}\\(r1024)}}      &   mAP(\%)               & 51.5    & 53.9   & 54.8   & 55.6 \\
                                &   Hours       & 38    &  75  & 197  & 402 \\
                     \midrule
				\multirow{2}{*}{\makecell[c]{Lite-SAM\\(r640)}}  &   mAP(\%)    & 52.3 & 55.8 & 56.4 & 57.1 \\
                                &   Hours       & 26 & 50   & 130  & 272 \\
                     \midrule
				\multirow{2}{*}{\makecell[c]{Lite-SAM\\(r1024)}} &   mAP(\%)   & 53.9  & 56.5  & 57.6 & 58.2  \\
                                &   Hours       & 37   & 68   & 181   & 403  \\                                                                          
				\bottomrule
			\end{tabular}
	}}
	
\end{table}

\begin{figure}[!ht]
	\centering
	\includegraphics[width=0.9\linewidth]{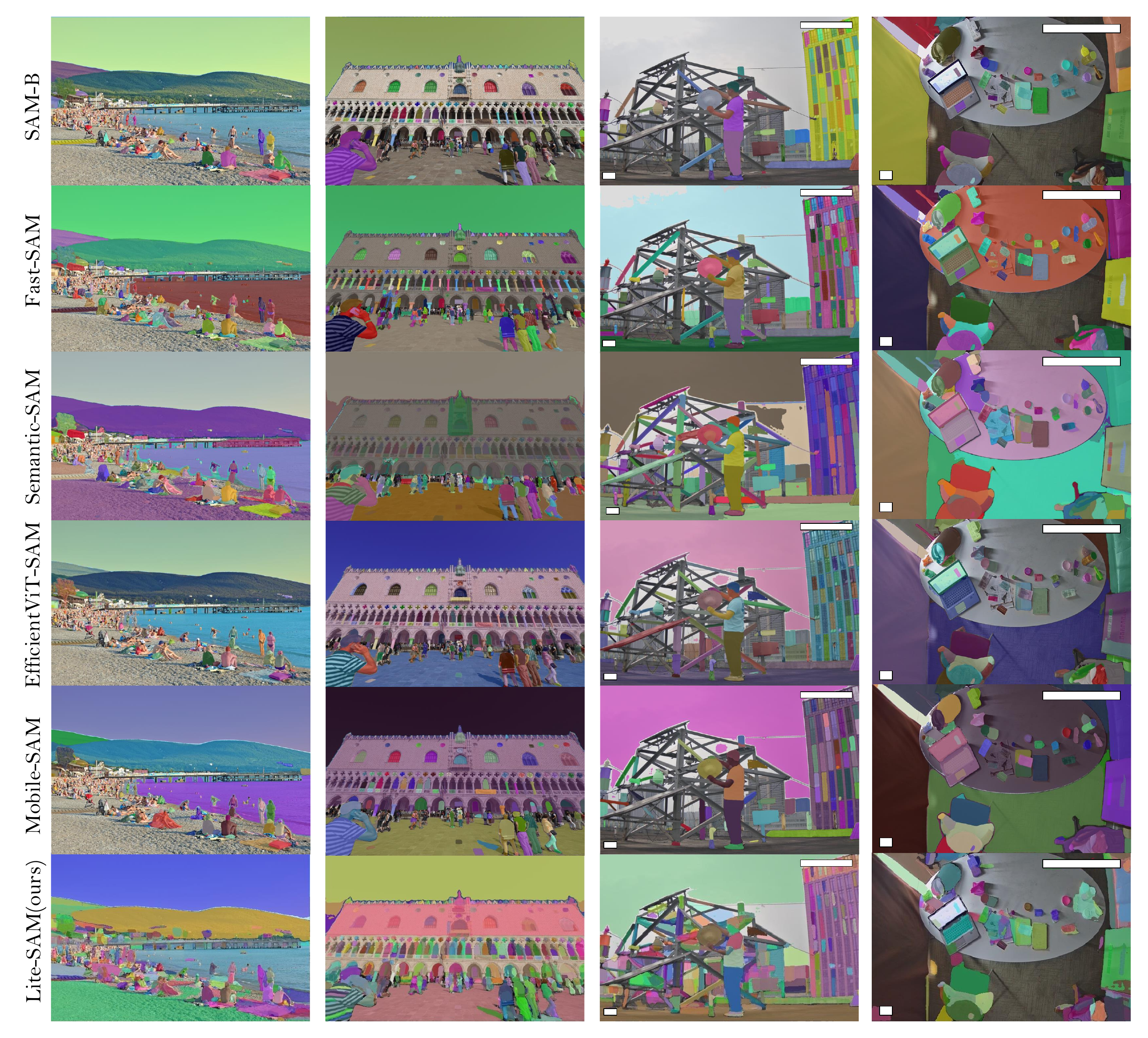}
	\caption{\textbf{Qualitative results on ``SegEvery''.} Models demonstrate mask generation capabilities. 
		(1) Note that EfficientViT-SAM's \cite{caiefficientvit} result is based on L1 model.
		(2) Lite-SAM employs an inference size of 640 $\times$ 640, while other comparison algorithms utilize a default size of 1024 $\times$ 1024.}
	\label{fig:everything}
\end{figure}

\subsection{Results for Segment Anything and Everything.}
The qualitative ``SegEvery'' outcomes of SAM\cite{kirillov2023segment} , Semantic-SAM \cite{li2023semantic},
Fast-SAM \cite{zhao2023fast}, Mobile-SAM \cite{zhang2023faster}, and EfficientSAM \cite{caiefficientvit}, 
and our proposed approach are depicted in \cref{fig:everything}.  
The visualization illustrates that Lite-SAM achieves comparable results to SAM-B \cite{kirillov2023segment} 
and exhibits superior performance over both Fast-SAM \cite{zhao2023fast} and Mobile-SAM \cite{zhang2023faster}.
We also provide ``SegAny'' visualized results and comparisions for box and point prompt in \cref{fig:prompts}.

\begin{figure}[!ht]
	\centering
	\includegraphics[width=0.9\linewidth]{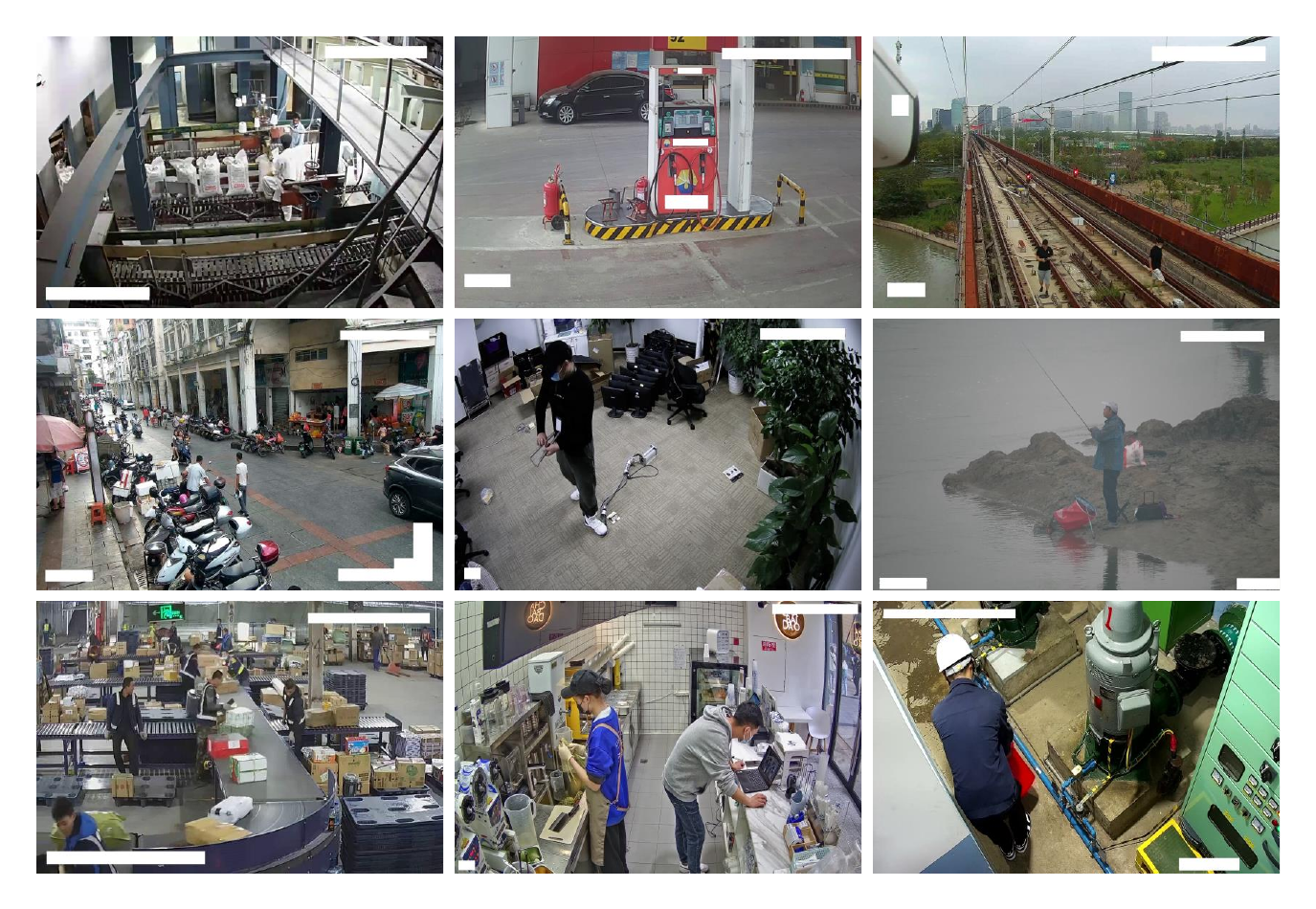}
	\caption{The proposed \textbf{ARI-TEST2024 dataset.} Faces and vehicle
		license plates have been blurred in the released images. 
		All images are resized to a size of 1024 $\times$ 1024. Each scene contains 1000 images, randomly selected from different videos.}
	\label{fig:ari2023}
\end{figure}

\subsection{Zero-Shot Image Segmentation Results on ARI-TEST2024}

\textbf{Private data.} To further evaluate the zero-shot generalization in real-world scenarios, we introduce a novel dataset termed \textbf{ARI-TEST2024}.
This dataset contains 10,000 meticulously annotated high-resolution images (1024 $\times$ 1024) from varied locations,
such as storage units, reservoirs, restaurant kitchens, transformer substations, gas stations, and garbage recycling facilities.
Representative samples are presented in \cref{fig:ari2023}.

\begin{figure}[!ht]
	\centering
	\includegraphics[width=0.9\linewidth]{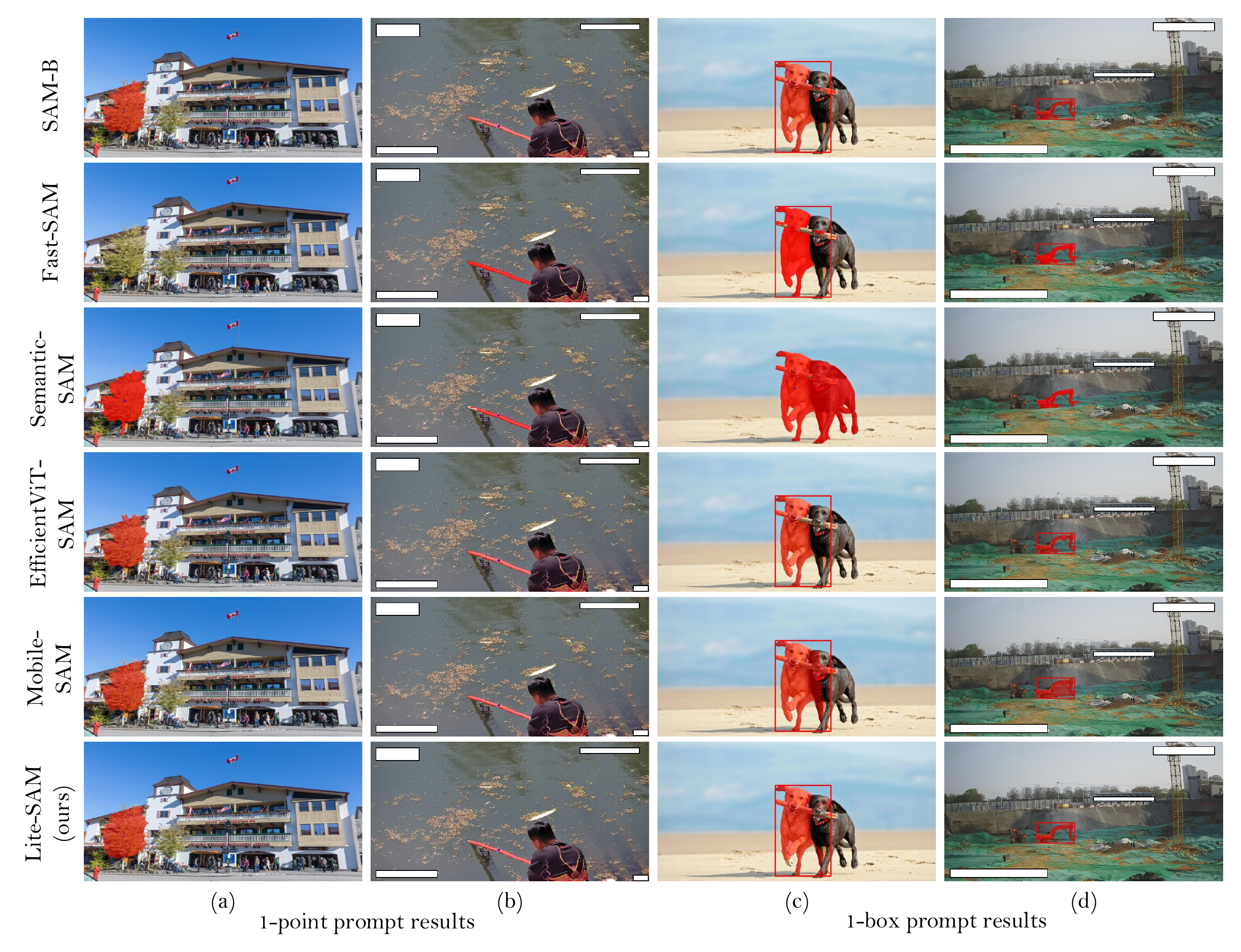}
	\caption{\textbf{Qualitative ``SegAny'' results on COCO2017 and ARI-TEST2024 with bounding box or point as prompt.}
		Please note that the code provided for Semantic-SAM \cite{li2023semantic} does not include support for box prompts. Therefore, we have used point prompt results instead. 
	}
	\label{fig:prompts}
\end{figure}

\begin{table}[!ht]
	\small\centering
     \caption{\textbf{Zero-Shot Image Segmentation Results on ARI-TEST2024 using mIoU metric.}}
    \label{tab:sam_ARI}
    \scalebox{0.9}
		{
	\begin{tabular}{l | c | c | c}
		\toprule
		\multirow{2}{*}{Model} & \multicolumn{3}{c}{ARI-TEST2024 mIoU $\uparrow$}  \\
		\cmidrule{2-4}   & 1-box($\%$) & 1-point($\%$) &  Input Size   \\
		\midrule
		SAM-B    \cite{kirillov2023segment}     & 70.6 & 53.7 & $1024^2$ \\
		SAM-L    \cite{kirillov2023segment}     & 72.3 & 56.4 & $1024^2$  \\
		SAM-H    \cite{kirillov2023segment}     & \textbf{72.4} & \textbf{56.8} & $1024^2$  \\
		\midrule
		Semantic-SAM \cite{li2023semantic}      & N/A & 50.3   & $1024^2$  \\
		Fast-SAM \cite{zhao2023fast}          & 62.2 & 41.5  & $1024^2$   \\
		Mobile-SAM \cite{zhang2023faster}     & 64.0 & 42.0  & $1024^2$  \\
		EfficientViT-L0-SAM \cite{caiefficientvit}    & 67.9 & 50.4  & $1024^2$  \\
		EfficientViT-L1-SAM \cite{caiefficientvit}    & \textbf{68.6} & 51.6  & $1024^2$  \\
		\midrule
		\textbf{Lite-SAM(r640)}                    & 66.6 & 52.1  & $640^2$  \\
		\textbf{Lite-SAM(r1024)}                   & 68.3 & \textbf{54.5}  & $1024^2$   \\
		\bottomrule
	\end{tabular}}	
\end{table}

We demonstrate the robust generalization and stability of our proposed Lite-SAM algorithm in comparison with eight different algorithms.
Lite-SAM achieves mIoU scores of 68.3\% and 54.5\% using the 1-box and 1-point prompt respectively, as detailed in \cref{tab:sam_ARI}.

To assess the effectiveness of our model in generating segmentation masks influenced by prompts, 
we utilize both our model and other models based on the SAM framework to conduct instance segmentation.
This includes both point-based and box-based prompt segmentation methodologies.
In \cref{fig:prompts}, it is evident that  Fast-SAM \cite{zhao2023fast} fails to produce any results in the scene shown in column (a).
This behavior can be attributed to the approach employed by the algorithm,
where the input point or box is treated solely as a post-processing strategy rather than being utilized as an actual cue.
In contrast, our Lite-SAM generates a satisfactory mask prediction that closely resembles the output obtained from SAM-B \cite{kirillov2023segment}.

\section{Other Materials}

\subsection{Distance Transform: pseudo code}

As mentioned in Section 3.3-(2) and Figure 4, 
we have incorporated the use of distance transforms to estimate the confidence of point prompts. 
This facilitates the calculation of 
the distance between a point and its corresponding mask, as depicted in \cref{code1}.

\begin{lstlisting}[language=Python, caption=distance transform pseudo code, label=code1]
alpha = 4.0; eps = 1e-4
h, w, c = image.shape; hmap = zeros(h, w)
masks = sort(masks, key=lambda x: x['area'], reverse=True) 
for mask in masks:
    mask_pad  = pad(mask, ((1, 1), (1, 1)), 'constant')
    mask_dist = distanceTransform(mask_pad, kernel=5)
    mask_dist = (mask_dist / (mask_dist.max() + eps))
    mask_dist = mask_dist ** alpha 
    hmap = maximum(hmap, mask_dist)
\end{lstlisting}

\subsection{Q \& A}

1. \textit{Why do the parameter and computation amounts differ from those mentioned in the reference article?}

Answer: The reference article does not provide the script to calculate the parameter and computation amounts. 
Hence, we downloaded their code and model, and employed the same script for an accurate calculation.
The script utilized is provided in \cref{code2}. (If the code was unavailable, we used the data provided in the reference article).

\begin{lstlisting}[language=Python, caption=complexity pseudo code, label=code2]
from ptflops import get_model_complexity_info as cmplx
from segment_anything import sam_model_registry

class Model_1prompt(torch.nn.Module):
  def __init__(self, model):
    super(FlopTestModel, self).__init__()
    self.model = model
    
  def forward(self, inputs):  
    image_embedding, _ = self.model.image_encoder(inputs)
      
    sparse_embeddings, dense_embeddings = \
        self.model.prompt_encoder(
          points=(torch.randn(1,1,2).cuda(), 
                  torch.randn(1,1).cuda()),
          boxes=None,
          masks=None,
        )
           
    low_res_masks, iou_predictions = \
        self.model.mask_decoder(
           image_embeddings=image_embedding,
           image_pe=self.model.prompt_encoder.get_dense_pe(),
           sparse_prompt_embeddings=sparse_embeddings,
           dense_prompt_embeddings=dense_embeddings,
           multimask_output=True,
        )       
    return low_res_masks, iou_predictions

if __name__ == "__main__":    
  model = sam_model_registry["vit_b"]
  model_1prompt = Model_1prompt(model)
  input_size = 640
  flops, params = cmplx(model.image_encoder, 
                        (3, input_size, input_size), 
                        as_strings=True, 
                        print_per_layer_stat=True)
  print("FLOPs: %s, Params: %s " % (flops, params))
  
  flops, params = cmplx(model_1prompt, 
                        (3, input_size, input_size), 
                        as_strings=True, 
                        print_per_layer_stat=False)
  print("FLOPs: %s, Params: %s " % (flops, params))    
\end{lstlisting}

2. \textit{Why is the inference time different from other papers?}

Answer: In this paper, we recalibrate the computation time of SegEvery, adopting the calculation method used by Mobile-SAM-v2~\cite{zhang2023mobilesamv2} for a uniform comparison.
The inference time mentioned in the SAM~\cite{kirillov2023segment} , Semantic-SAM~\cite{li2023semantic}, Fast-SAM~\cite{zhao2023fast}, 
EfficientViT-SAM~\cite{caiefficientvit} , Mobile-SAM~\cite{zhang2023faster}, and Edge-SAM~\cite{zhou2023edgesam} papers refers to the SegAny time.
However, the inference time reported in this paper and Mobile-SAM-v2~\cite{zhang2023mobilesamv2} is based on SegEvery time. Additionally, Mobile-SAM-v2~\cite{zhang2023mobilesamv2}
is a two-stage model, and the parameter count and inference time of the Object-aware model are not reported in the paper. 
Therefore, we have recalculated the parameter count, MACs, and SegEvery time for Mobile-SAM-v2~\cite{zhang2023mobilesamv2}.

3. \textit{Does this paper solely support segmentation? Does it have text capabilities?}

Answer: In this paper, benchmarking against lightweight SAM algorithms like MobileSAM~\cite{zhang2023faster} and Mobile-SAM-v2~\cite{zhang2023mobilesamv2}, 
primarily addresses the SegEvery problem. It does not support text capabilities.

\end{document}